\documentclass{article}

\usepackage[main, final]{neurips_2025}

\usepackage[utf8]{inputenc} 
\usepackage[T1]{fontenc}    
\usepackage{hyperref}       
\usepackage{url}            
\usepackage{booktabs}       
\usepackage{amsfonts}       
\usepackage{nicefrac}       
\usepackage{microtype}      
\usepackage{xcolor}         
\usepackage{mdframed}
\usepackage{color}
\usepackage{xcolor}
\usepackage[utf8]{inputenc} 
\usepackage[T1]{fontenc}    

\usepackage{amsfonts}       
\usepackage{nicefrac}       
\usepackage{microtype}      
\usepackage{multirow}
\usepackage{multicol}
\usepackage{tabto}
\usepackage{xspace}
\usepackage{amsmath}
\usepackage{adjustbox}
\usepackage{enumitem}
\usepackage{wrapfig}
\usepackage{dblfloatfix}
\usepackage{times}
\usepackage{verbatim}
\usepackage{amssymb}
\usepackage{mathtools}
\usepackage{caption}
\usepackage{subcaption}
\usepackage{array}
\usepackage{colortbl}
\usepackage{booktabs}
\usepackage{bbm}
\usepackage{makecell}
\usepackage{float}
\usepackage{siunitx}
\usepackage{pifont}
\usepackage{marvosym}
\usepackage{listings}
\usepackage{pdflscape}
\usepackage{footmisc}
\usepackage{url}
\usepackage{tabularx}
\usepackage{arydshln}
\usepackage{hhline}
\usepackage{diagbox}
\usepackage{tcolorbox}
\usepackage[nameinlink]{cleveref}
\usepackage{hyperref}
\usepackage{fp} 
\usepackage{authblk}
\usepackage{xspace}
\newcommand{\cmark}{\ding{51}}  
\newcommand{\xmark}{\ding{55}}  

\title{Scaling RL to Long Videos}

\author{%
\begin{minipage}[t]{\textwidth}
\centering
Yukang Chen$^{1*}$\quad 
Wei Huang$^{1,3*}$ \quad 
Baifeng Shi$^{1,4\dagger}$ \quad 
Qinghao Hu$^{2\dagger}$  \quad 
Hanrong Ye$^{1\dagger}$ \quad \protect\\[0.1cm]
Ligeng Zhu$^{1}$ \quad 
Zhijian Liu$^{1}$ \quad 
Pavlo Molchanov$^{1}$ \quad 
Jan Kautz$^{1}$ \quad 
Xiaojuan Qi$^{3}$ \quad \protect\\[0.1cm]
Sifei Liu$^{1}$ \quad 
Hongxu Yin$^{1}$ \quad 
Yao Lu$^{1}$ \quad 
Song Han$^{1,2}$ \protect\\[0.2cm]
$^1$NVIDIA \quad $^2$MIT \quad $^3$HKU \quad $^4$UC Berkeley
\end{minipage}
}

\begin{document}

\maketitle
{\renewcommand\thefootnote{\fnsymbol{footnote}}\footnotetext[0]{$^*$Equal contribution ~~ $^\dagger$Core contribution}}

\begin{abstract}
We introduce a full-stack framework that scales up reasoning in vision-language models (VLMs) to long videos, leveraging reinforcement learning.
We address the unique challenges of long video reasoning by integrating three critical components: (1) a large-scale dataset, LongVideo-Reason, comprising 104K long video QA pairs with high-quality reasoning annotations across diverse domains such as sports, games, and vlogs; (2) a two-stage training pipeline that extends VLMs with chain-of-thought supervised fine-tuning (CoT-SFT) and reinforcement learning (RL); and (3) a training infrastructure for long video RL, named Multi-modal Reinforcement Sequence Parallelism (MR-SP), which incorporates sequence parallelism and a vLLM-based engine tailored for long video, using cached video embeddings for efficient rollout and prefilling.
In our experiments, LongVILA-R1-7B achieves strong performance on video benchmarks, reaching 65.1\% and 71.1\% accuracy on VideoMME without and with subtitles, respectively, and consistently outperforming LongVILA-7B across multiple benchmarks. Moreover, LongVILA-R1-7B supports processing up to 8,192 video frames per video, and configurable FPS settings. Notably, our MR-SP system achieves up to 2.1$\times$ speedup on long video RL training. In addition, we release our training system for public availability that supports RL training on various modalities (video, text, and audio), various models (VILA and Qwen series), and even image and video generation models. On a single A100 node (8 GPUs), it supports RL training on hour-long videos ({\em e.g.}, 3,600 frames). Code and models are available at 
\href{https://github.com/NVlabs/Long-RL}{https://github.com/NVlabs/Long-RL} and \href{https://huggingface.co/Efficient-Large-Model/LongVILA-R1-7B}{https://huggingface.co/Efficient-Large-Model/LongVILA-R1-7B}.
\end{abstract}

\section{Introduction}
\begin{figure*}[!t]
\centerline{\includegraphics[width=1.0\textwidth]{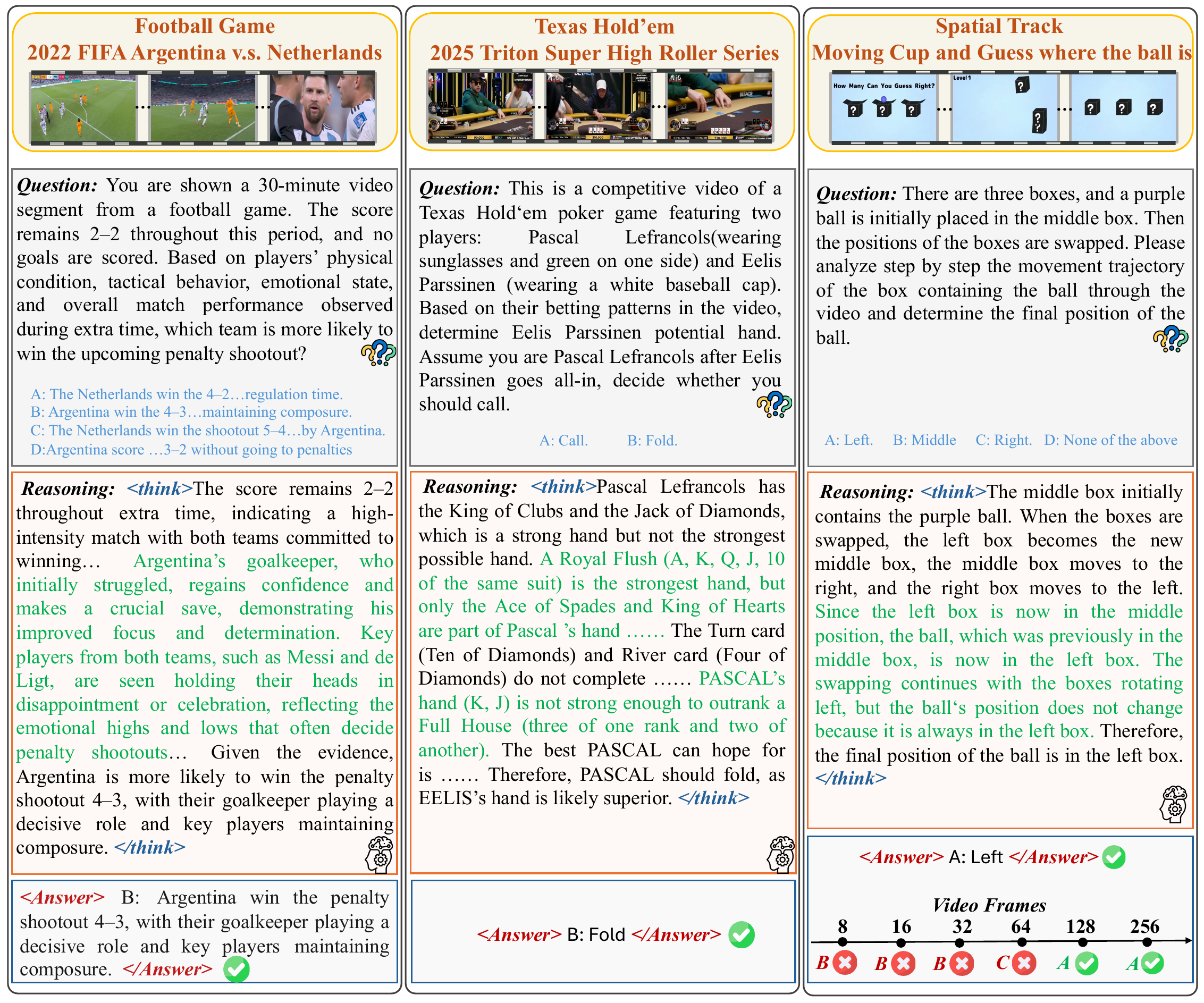}}
\caption{Examples of LongVILA-R1. The illustration demonstrates sample tasks and their reasoning. From left to right, the examples include predicting the results of a football match, decision-making reasoning in Texas Hold'em Poker, and trajectory for spatial dynamics of objects. Notably, the spatial tracking video involves a relatively complex dynamic moving, for which the model fails to achieve accurate reasoning until the number of input video frames increases to 128.}
\label{fig:examples}
\end{figure*}

Understanding long videos requires more than simple recognition—it demands reasoning from temporal, spatial, goal-oriented, and narrative perspectives~\cite{zhang2024long}. As illustrated in Figure~\ref{fig:examples}, answering high-level questions often hinges on a model’s ability to integrate clues distributed across time, infer hidden goals or strategies, track entities spatially, and comprehend the evolving plot. For instance, predicting the winner of a football penalty shootout involves assessing emotional cues and tactical behavior (temporal and goal reasoning), while determining the final location of a hidden ball requires precise spatial tracking. 
Likewise, evaluating a poker player’s decision demands interpreting implicit strategies beyond surface actions (goal reasoning) and understanding a character’s development or match trajectory reflects the need for plot reasoning. These examples underscore that reasoning is indispensable for long video understanding that goes beyond recognition alone.
Despite the clear importance of reasoning in long video understanding, enabling such capabilities in long video VLMs poses significant challenges~\cite{feng2025video,weng2024longvlm,chen2024longvila,guo2025fila,shen2025long}. First, the collection of high quality long video reasoning datasets is inherently difficult. Unlike domains such as math or code reasoning, where structured supervision and benchmarks are readily available~\cite{shao2024deepseekmath,liu2024codemind}, long video reasoning requires annotating complex temporal dynamics, goals, spatial relations, and narrative elements—often across minutes or hours of footage~\cite{han2024videoespresso}. This process is labor-intensive and subjective, making large-scale dataset construction slow and costly. Second, the RL training framework for long videos is challenging. Reinforcement learning, a common strategy for aligning models with complex reasoning objectives, is computationally expensive and sample-inefficient~\cite{sheng2024hybridflow}. When applied to long videos, RL becomes even more burdensome due to the extended video frames, requiring more memory and longer rollout runtime. These challenges jointly hinder the development of effective long video VLMs with strong reasoning capabilities.

In this work, we introduce LongVILA-R1, a comprehensive framework exploring the reasoning capabilities for long video understandings. Firstly, we strategically construct a high quality dataset with CoT annotations for long video reasoning, named LongVideo-Reason. Leveraging a powerful VLM (NVILA-8B)~\cite{liu2024nvila} and a leading open-source reasoning LLM, we develop a dataset comprising 104K high quality \textit{Question-Reasoning-Answer} pairs for long videos. We use 36K high quality samples for Long-CoT-SFT to initialize the model's reasoning and instruction-following abilities, and 68K samples with an additional 102K video data~\cite{llava-video,next-qa,perpceptiontest,clevr,star} for reinforcement learning. This two-stage training combines high quality reasoning annotations with reinforcement learning, enabling LongVILA-R1 to achieve superior and generalized video reasoning. We also manually curate a balanced set of 1K long video samples to build a new benchmark, LongVideo-Reason-eval, that evaluates performance from four perspectives: \textit{Temporal}, \textit{Goal and Purpose}, \textit{Spatial}, and \textit{Plot and Narrative}, for a comprehensive assessment.

\begin{figure}[!ht]
    \centering 
    \includegraphics[width=\columnwidth]{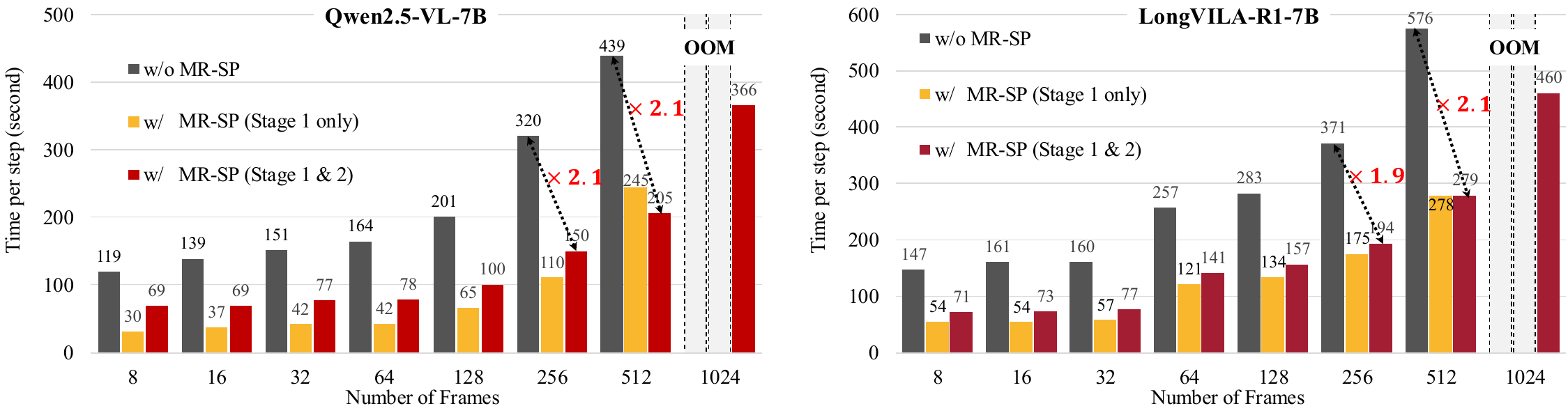}
  \caption{Training efficiency comparison with MR-SP (SP degree=4) on Qwen2.5-VL-7B and LongVILA-R1-7B and a single node 8$\times$ A100 GPUs. It achieves 2.1$\times$ speed-up and avoids GPU OOM issue on long frames.}
    \label{fig:efficiency-mrsp}
\end{figure}

Secondly, we propose a training framework for VLMs to advance long video reasoning. As illustrated in Figure~\ref{fig:training-pipeline}, this framework incorporates two stages, {\em i.e.}, Stage-1: Long CoT-SFT, and Stage-2: RL for long video reasoning. To address the unique challenges of long video RL, including massive visual embeddings, heavy rollouts, and long-context LLM prefilling, we develop an efficient and scalable solution, referred to as Multi-modal Reinforcement Sequence Parallelism (MR-SP). It incorporates a vLLM engine~\cite{kwon2023efficient} tailored for LongVILA and a caching scheme for video embeddings. The MR-SP system alleviates the problem of intensive memory and facilitates RL training of long video VLMs. As shown in Figure~\ref{fig:efficiency-mrsp}, our MR-SP system achieves up to 2.1$\times$ speedup on 512-frame video RL training on 7B models and enables longer training frames without out-of-memory (OOM).

In our experiments, LongVILA-R1-7B demonstrates strong performance on video benchmarks, achieving 65.1\% and 71.1\% accuracy on VideoMME without and with subtitles, respectively. It consistently outperforms LongVILA-7B across a range of benchmarks, including ActivityNet-QA, LongVideoBench, PerceptionTest, NExT-QA, VNBench, and VideoMME.
Additionally, LongVILA-R1 supports processing up to 8,192 video frames per video, and configurable FPS settings.
On our LongVideo-Reason-eval benchmark, LongVILA‑R1‑7B achieves an average accuracy of 72.0\%, surpassing Video‑R1‑7B\cite{feng2025video}, as well as proprietary models such as Gemini‑1.5‑Pro\cite{team2024gemini}.
\section{Related Work}
\paragraph{Multi-modal reasoning models.} 
The field of multi-modal reasoning has advanced significantly, particularly in Vision-Language Models (VLMs). GPT-4o~\cite{gpt4o} improves visual understanding through enhanced reasoning, while Gemini-1.5-Pro~\cite{team2024gemini} extends context length to 1 million tokens, achieving state-of-the-art performance in VideoMME~\cite{video-mme}. Following the substantial progress of architecture and training algorithms in open-source VLMs~\cite{zhu2025internvl3,liu2024nvila,yang2024qwen2,lin2024vila,vila-hd}, multi-modal reasoning has been further explored in works including LMM-R1~\cite{peng2025lmm} which employs a two-stage training strategy, Vision-R1~\cite{huang2025vision} that addresses post-cold-start overthinking, and Video-R1~\cite{feng2025video} which enhances RL for video via T-GRPO in 16 frames. However, these approaches primarily focus on single images or short videos, and long video reasoning still poses great challenges.
\begin{figure*}[!t]
\centerline{\includegraphics[width=1\textwidth]{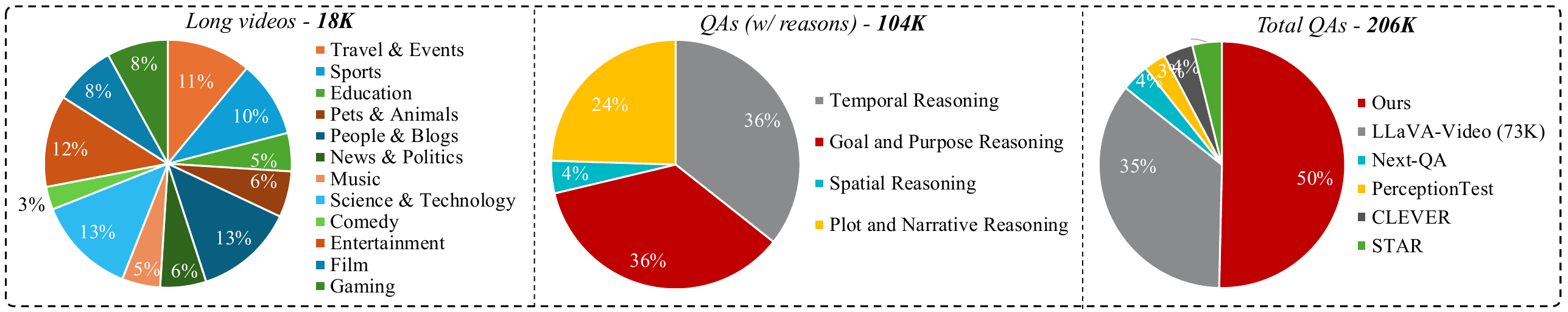}}
\caption{Data Distribution of LongVideo-Reason and total data in the LongVILA-R1 training framework. LongVideo-Reason comprises a total of 18K videos and 104K QAs with reasoning annotations. Additionally, we include 102K QAs from existing works~\cite{llava-video,next-qa,perpceptiontest,clevr,star}.}
\label{fig:data-distribution}
\end{figure*}

\paragraph{Sequence parallelism.} 
Training with long contexts often exceeds the memory capacity of a single device~\cite{longlora}, necessitating efficient distribution strategies. Sequence parallelism (SP) has become a widely adopted solution ~\cite{liu2023ring, jacobs2023deepspeed, InternEvo, USP, LoongTrain}. For example, ring-based systems like LightSeq~\cite{li2023lightseq} and Ring Attention~\cite{liu2023ring} use point-to-point (P2P) communication, while DeepSpeed-Ulysses~\cite{jacobs2023deepspeed} employs all-to-all (A2A) primitives to optimize attention computations. Additionally, USP and LoongTrain ~\cite{USP, LoongTrain} were introduced to integrate Ring-style SP and Ulysses SP. LongVILA~\cite{chen2024longvila} further proposed multi-modal SP (MM-SP), enabling vision-language models to handle long-context inputs. However, multi-modal reinforcement learning introduces additional challenges, as it requires extensive sampling from long, mixed-token sequences~\cite{sheng2024hybridflow}, particularly in complex group optimization tasks~\cite{shao2024deepseekmath}.

\paragraph{RL frameworks for LLMs/VLMs.} 
Reinforcement Learning (RL) has become a key strategy for enhancing Large Language Models (LLMs), particularly through Reinforcement Learning with Human Feedback (RLHF)~\cite{ouyang2022training} or Direct Preference Optimization~\cite{dpo}, which aligns model outputs with human preferences.  Recent advancements demonstrate that RL significantly improves LLM reasoning abilities. For example, DeepSeek-R1~\cite{guo2025deepseek} utilizes the Group Relative Policy Optimization (GRPO) algorithm~\cite{grpo}, integrating group-based sampling and rule-based rewards. On the other hand, RL~\cite{chen2025r1v} poses a unique challenge of heavy computational cost, especially in multi-modal settings. To address this problem, HybridFlow~\cite{sheng2024hybridflow} is introduced, leveraging Ray~\cite{moritz2018ray} for efficient data flow and vLLM~\cite{kwon2023efficient} for faster sampling. Nevertheless, this remains a bottleneck when processing long video sequences, with group-based sampling constrained by the high computational cost of long-context sampling and visual encoding. In this work, we propose MR-SP that provides up to 2.1$\times$ speedup.

\section{LongVideo-Reason Data Construction}

\subsection{Overview of Data Curation}\label{sec:data_curation}
We first curate 18K long videos from the Shot2Story dataset~\cite{han2023shot2story20k} (Figure~\ref{fig:data-distribution}, left). We further incorporated 2k additional 4K-resolution videos spanning scenarios such as autonomous driving, video games, household robotics, and wildlife. We then apply a high quality automated annotation pipeline for CoT as detailed in Section~\ref{sec:cot_generation}, and end up with a total of 104K Question-Reasoning-Answer pairs where each sample, based on the type of question it is reasoning about, can be categorized into \textit{Temporal Reasoning}, \textit{Goal and Purpose Reasoning}, \textit{Spatial Reasoning}, or \textit{Plot and Narrative Reasoning} (Figure~\ref{fig:data-distribution}, middle). This dataset is designed to support various types of long-video reasoning tasks comprehensively.

Given the sensitivity of GRPO to batch sampling~\cite{team2025kimi,peng2025lmm}, a data filtering approach is adopted. Specifically, we employ a test-scaling method where LongVILA~\cite{chen2024longvila} performs inference 10 times on the original datasets. Questions consistently answered correctly or incorrectly are labeled as easy or hard while those inducing diverse predictions are labeled as medium. 
We filter out too easy and too hard questions, and use remaining questions for training. The reason is that GRPO expects different rollouts of each sample to be diverse in order to have meaningful advantages, and the gradient vanishes if all the rollouts predict correct or incorrect answers. The COT-SFT subset (36K) features high quality CoT reasoning processes formatted in a standard \textit{ <think></think><answer></answer>} structure, providing abundant resources for warm-up training during Stage 1 of the model's reasoning capabilities. Meanwhile, the RL subset contains 68K challenging long-video Q\&A, which is leveraged in Stage 2 for scaling reasoning through reinforcement learning. To further enhance RL scaling, we incorporate an additional 102K high quality open-source videos (Figure~\ref{fig:data-distribution}, right) from other datasets~\cite{llava-video,next-qa,perpceptiontest,clevr,star}. This combination improves the model's generalization.

\begin{figure*}[t]
\centerline{\includegraphics[width=1\textwidth]{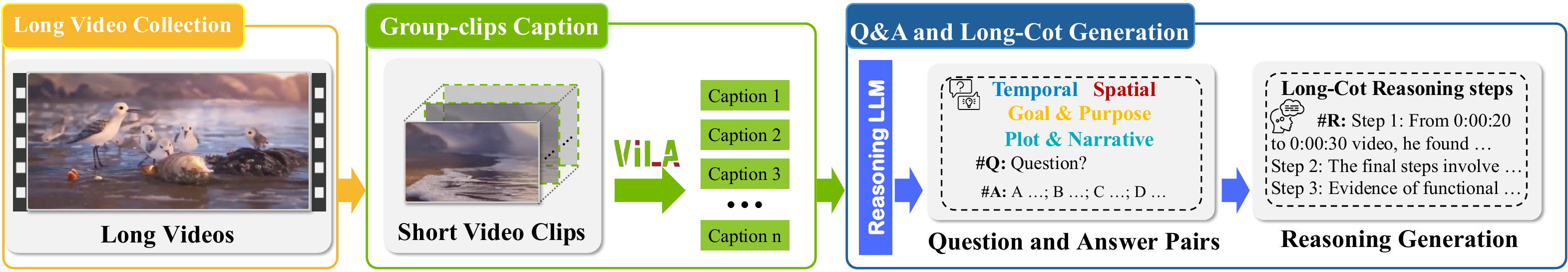}}
\caption{Data generation process for the LongVideo-Reason dataset. This process begins with segmenting videos into 10-second clips and generating captions for each clip using NVILA-8B. Then based on the captions of all clips in a video, we generate question-answer pairs that involve reasoning across the content of the whole video, along with the reasoning annotations using a leading open-source reasoning LLM. Reasoning questions are categorized into Temporal, Goal and Purpose, Spatial, and Plot and Narrative. Finally, the reasoning annotations are reformatted for conciseness and alignment with video details. We present a more detailed figure of data generation process in Figure~\ref{fig:data-generation-appendix}.
}
\label{fig:data-generation}
\end{figure*}

\subsection{Long-Video Reasoning Generation}
\label{sec:cot_generation}
We introduce an automated annotation pipeline (Figure~\ref{fig:data-generation}) that generates high quality \textit{Question-Reasoning-Answer} pairs from long videos. This process begins by segmenting videos into short clips (\textasciitilde 10 seconds each), each of which is annotated using the NVILA-8B~\cite{liu2024nvila} model to provide descriptive captions. For the spatial reasoning category, We employed VILA-HD~\cite{vila-hd} to generate object bounding boxes in video frames and constructed spatial reasoning QAs based on these boxes and corresponding captions. Leveraging the breakthrough in text-based reasoning, we then deploy a leading open-source reasoning LLM, DeepSeek-R1-671B~\cite{guo2025deepseek}, provide the captions of all the clips in each video, and prompt it to generate diverse types of \textit{Question-Reasoning-Answer} pairs that involve reasoning over the content across the whole video. 

Specifically, we design four types of prompts to encourage the LLM to generate a \textit{Question-Reasoning-Answer} pair that focuses on one of the four types of reasoning: \textit{Temporal Reasoning}, \textit{Goal and Purpose Reasoning}, \textit{Spatial Reasoning}, or \textit{Plot and Narrative Reasoning}. To ensure VLMs focus on visual details, we also craft the prompts with phrases such as \textit{“checking the video”} and \textit{“analyzing the scene”}, which guide iterative examination of visual content. Finally, an LLM is then used to refine and streamline the reasoning steps. We also manually curated 1,000 high quality complex reasoning questions across four reasoning categories to serve as a new benchmark (LongVideo-Reason-eval) for evaluating VLMs in reasoning abilities. 
Note that we also include open-ended questions. Among the total 104K QA pairs, approximately half are multiple-choice and hald are open-ended. This entire data procedure consumes about 80,000 H100 GPU hours. 

\section{LongVILA Training Pipeline}
\begin{figure*}[!t]
\centerline{\includegraphics[width=1\textwidth]{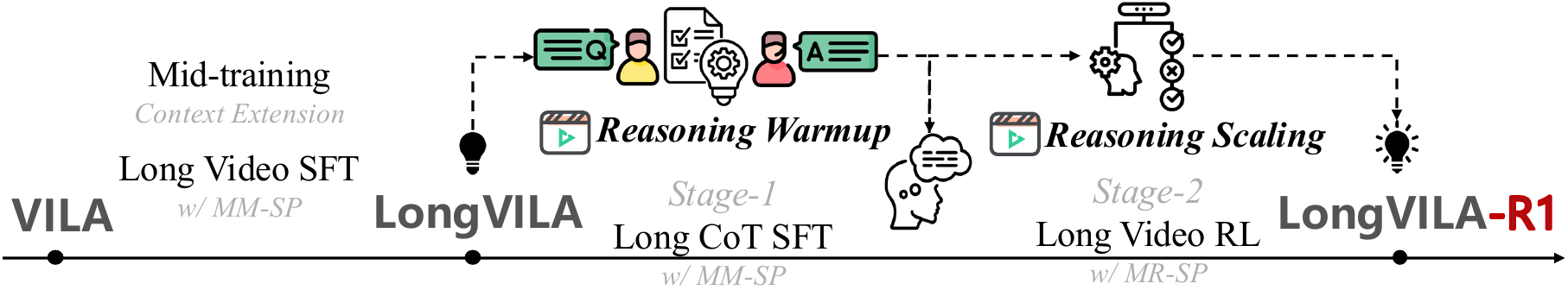}}
\caption{The LongVILA-R1 training pipeline. LongVILA-R1 builds upon the base training pipeline for LongVILA. MM-SP is further employed for SFT on long video understanding tasks with long CoT. Then, reinforcement scaling learning is conducted through Multi-modal Reinforcement Sequential Parallelism (MR-SP).}
\label{fig:training-pipeline}
\end{figure*}

As shown in Figure~\ref{fig:training-pipeline}, there are two extended training stages in LongVILA-R1, \textit{i.e.}, (1) warm-up for long video reasoning, utilizing 36K data with high quality CoT for SFT on the MM-SP system~\cite{chen2024longvila}; (2) reinforcement learning with dense frames from long videos. \subsection{Long Video CoT Supervised Fine-Tuning}
Utilizing 104K high quality question-reasoning-answer pairs, we apply the data filtering method described in Section~\ref{sec:data_curation} to select 36K examples for long CoT-SFT, serving as a warm-up phase for subsequent RL. This stage equips the model with fundamental reasoning abilities and instruction-following skills for long video scenarios. To efficiently perform SFT on hundreds of frames, we adopt the MM-SP~\cite{chen2024longvila} training system from LongVILA. As demonstrated in Section~\ref{sec:ablations}, SFT solely on our LongVideo-Reason dataset also effectively improves the model with reasoning capabilities.


\subsection{GRPO for Long Video}
Building on the advancements of the GRPO~\cite{shao2024deepseekmath} algorithm and prior explorations of multi-modal reasoning training~\cite{feng2025video,peng2025lmm}, we adhere to the standard GRPO framework to train our model. For each given question $q$, the policy model generates a group of candidate responses $\{o_1, o_2,...,o_G\}$ from the old policy $\pi_{\theta_{old}}$, accompanied by their corresponding rewards $\{r_1, r_2,...,r_G\}$, which are computed based on rule-based reward functions (format/accuracy). The model $\pi_{\theta}$ is subsequently optimized by maximizing the following objective function:
\begin{multline}\label{eq1:grpo}
\mathcal{J}(\theta) = \mathbb{E}_{q,\{o_i\}} [ \frac{1}{G} \sum_{i=1}^{G} (\mathrm{min}(\frac{\pi_{\theta}(o_i|q)}{\pi_{\theta_{old}}(o_i|q)}A_i, \\ \mathrm{clip}(\frac{\pi_{\theta}(o_i|q)}{\pi_{\theta_{old}}(o_i|q)}, 1-\epsilon, 1+\epsilon)A_i)- \beta\mathbb{D}_{KL}(\pi_\theta||\pi_{ref}))]
\end{multline}
where $\epsilon$ and $\beta$ are hyper-parameters, $G$ is set as 8 in our experiments, and the sampled rewards above are normalized to get the advantages ($A_i$) for updating the model:

\begin{equation}\label{eq2:grpo}
A_i = \frac{r_i - \mathrm{mean}(\{r_1, r_2,...,r_G\})}{\mathrm{std}(\{r_1, r_2,...,r_G\})}
\end{equation}

\begin{figure*}[!t]
\centerline{\includegraphics[width=1\textwidth]{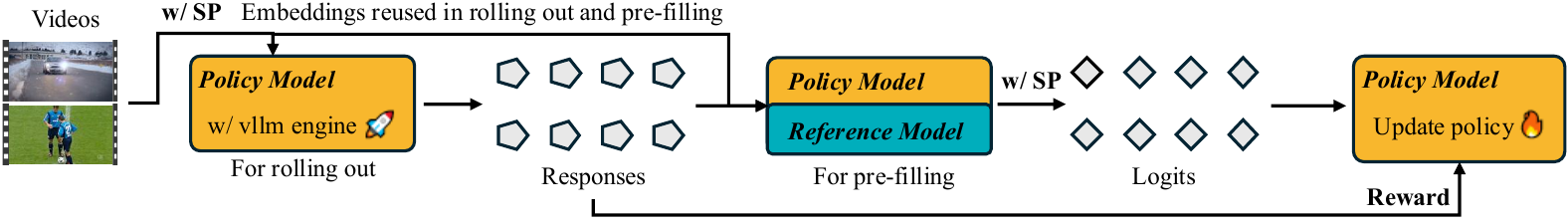}}
\caption{LongVILA-R1 RL training framework. The framework integrates multi-modal reinforcement sequence parallelism (MR-SP) for scalable video frame encoding and LLM prefilling. RL utilizes a vLLM-based engine with cached video embeddings, tailored for LongVILA rollout. Rewards for accuracy and format guide policy optimization.}
\label{fig:training-framework}
\end{figure*}
However, RL for long videos presents significant challenges due to the high computational demands of processing hundreds to thousands of frames. Existing RL frameworks struggle with such long-context training in rollout and LLM prefilling. To address this, we develop the MR-SP framework (Section~\ref{sec:4.2}), which efficiently scales reinforcement learning for long-context video reasoning.

Considering the sensitivity of GRPO to sampling during training~\cite{team2025kimi}, we use the 68K filtered data for reinforcement learning as described in Section~\ref{sec:data_curation}. Additionally, an extra 102K samples from other datasets~\cite{llava-video,next-qa,perpceptiontest,clevr,star} are incorporated to scale up the RL. This approach aims to guide the model in freely exploring and developing more effective and generalized reasoning strategies.

\section{Multi-modal Reinforcement SP}\label{sec:4.2}
\begin{figure*}[t]
    \centering
    \includegraphics[width=\linewidth]{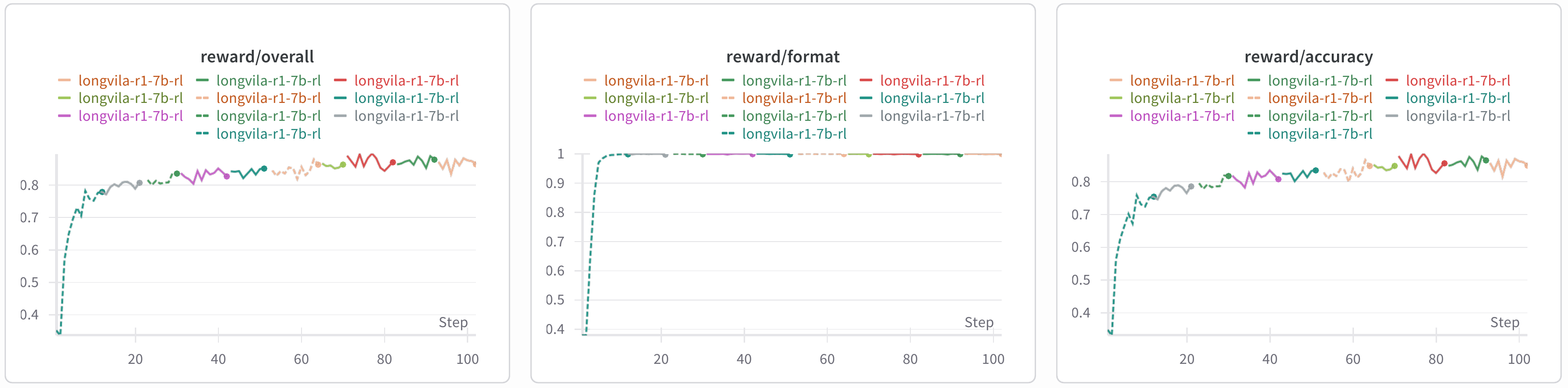}
    \caption{Reward curve of our LongVILA-R1-7B with MR-SP. During training, LongVILA-R1-7B exhibits stable reward improvements across overall, format, and accuracy dimensions.}
    \label{fig:reward-curce}
\end{figure*}

Existing RL frameworks for VLMs, such as R1-V~\cite{chen2025r1v} and EasyR1 \cite{EasyR1}, are not designed for long videos which present unique challenges due to their high token volume. To address this, we introduce Multi-modal Reinforcement Sequence Parallelism (MR-SP), a framework for efficient RL training on long videos. MR-SP leverages sequence parallelism in both rollout and pre-filling stages, enabling long videos in RL, with reduced overhead. We show the training curve with MR-SP in Figure~\ref{fig:reward-curce}.

\subsection{Stage 1 - Rollout with Paralleled Encoding}
\begin{figure*}[!t]
\centerline{\includegraphics[width=\textwidth]{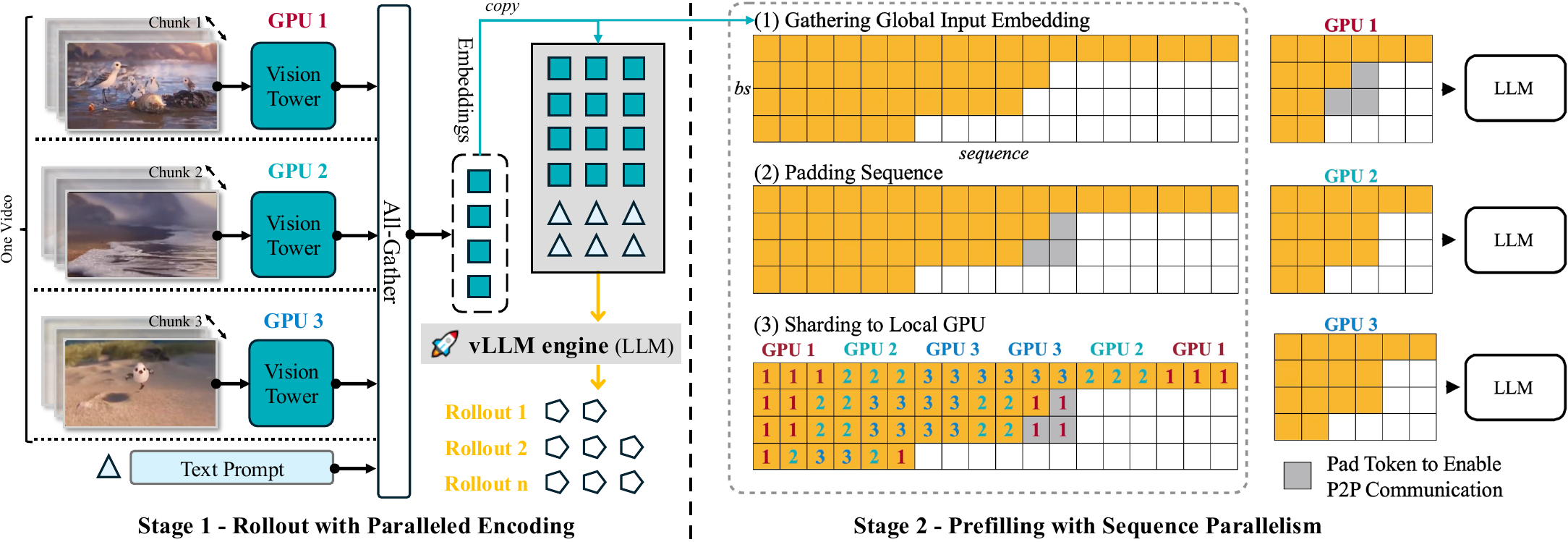}}
\caption{The workflow of multi-modal reinforcement sequential parallel (MR-SP). To accommodate multi-modal input in reinforcement learning, we develop a custom sharding strategy to ensure balanced workload distribution and compatibility with SP communication. Efficient video embedding reuse and vLLM rollout acceleration strategies are implemented, while meeting the demands of policy model prefilling for dense video frames.}
\label{fig:mrsp}
\end{figure*}

To support long-video reinforcement learning efficiently, we adopt sequence parallelism (SP) for the video encoding stage. As shown in Figure~\ref{fig:mrsp}, the input video frames are first evenly divided across multiple GPUs (e.g., GPU 1 to GPU 3), each equipped with its own vision tower. Each GPU independently processes a slice of the video, encoding only a subset of the frames. The resulting video embeddings are then aggregated with text embeddings via an all-gather operation as indicated by the "All-Gather" arrow in the figure. This strategy distributes the encoding workload, allowing the system to handle significantly longer videos by leveraging more GPUs, while avoiding the risk of GPU memory overflow. The parallel encoding scheme ensures balanced utilization of the vision towers and enables scalable long-video processing that would otherwise be infeasible.

After the video embeddings are globally gathered, they are \textbf{reused} for downstream usage throughout the RL pipeline. As illustrated in Figure~\ref{fig:mrsp}, the gathered embeddings are reused during multiple rollouts without recomputation. For instance, in each training step, we typically perform 8 to 16 rollouts. Without recycling, the same video would need to be re-encoded dozens of times per step, severely impacting training speed. By caching and reusing the gathered embeddings, MR-SP eliminates this redundancy and significantly accelerates training. 


\subsection{Stage 2 - Prefilling with Sequence Parallelism}
For each rollout, both the reference and policy models require prefilling compute-intensively in RL for long videos.
With the gathered embedding from Stage~1 \textbf{reused}, we parallelize the inference stage across devices using sequence parallelism. As illustrated in Figure~\ref{fig:mrsp}, we globally gathered input embeddings are first padded to a uniform length (Padding Sequence) and then evenly partitioned across GPUs (Sharding to Local GPU). This allows each GPU to handle only a portion of the input sequence during prefilling. This parallelism is applied to both policy and reference model prefilling. Then, each GPU locally computes logits for its token slice, prefilling in parallel.

\section{Experimental Results}
\subsection{Main Results}\label{sec:main_results}
Table~\ref{tab:9-benchmarks} shows the performance comparison on 6 video benchmarks~\citep{activity-qa,longvideobench,perpceptiontest,next-qa,vnbench,video-mme}. LongVILA-R1-7B consistently outperforms LongVILA-7B across all benchmarks, with performance gaps that vary according to the complexity of the reasoning tasks.
Table~\ref{tab:video-mme} presents the general performance of LongVILA-R1, comparing to existing advanced models~\cite{lin2023video,zhang2024beyond,chen2024sharegpt4video,li2023videochat, jin2024chat,liu2024kangaroo,wang2024retake,zhang2024long,feng2025video,apollo,qwen2-vl,internvl2.5,llava-video,liu2024nvila,videollama3} under comparable model sizes on the Video-MME~\cite{video-mme} benchmark. The LongVILA-R1-7B is tested using 512 video frames as inputs. LongVILA-R1-7B achieves leading scores across different video lengths, obtaining scores of 65.1\% and 71.1\% in the settings without subtitles and with subtitles, respectively.
Table~\ref{tab:LongVideo-Reason} compares the results of our LongVideo-Reason-eval benchmark. LongVILA-R1-7B model achieves a strong performance with an average score of 72.0\%, surpassing Video-R1-7B~\cite{feng2025video} and slightly outperforms Gemini-1.5-Pro~\cite{team2024gemini}. Qualitative results are in the appendix.

\subsection{Ablation Study}
\label{sec:ablations}
\paragraph{Scaling video frames.}
The reasoning capability of LongVILA-R1 scales consistently with the number of input video frames. Specifically, Table~\ref{tab:scaling-video-frames} illustrates the performance of LongVILA-1.5B (grey line) and LongVILA-1.5B-R1 (red line) on the long-video reasoning benchmark under varying frame inputs. With only 16 input frames, LongVILA-R1-1.5B is close to LongVILA-1.5B. However, as the number of frames increases to 512, LongVILA-R1-1.5B consistently outperforms and eventually achieves a score of 64.3\%. Notably, LongVILA-R1-1.5B demonstrates steady performance improvements throughout the scaling process. In contrast, LongVILA-1.5B hits a performance bottleneck with 256 frames, and got a degradation on 512 frames. The enhanced reasoning capabilities of LongVILA-R1-1.5B allow it to effectively integrate and infer information from long videos.

\begin{table*}[t]
\begin{center}
\caption{{{
Performance on ActivityNet-QA~\citep{activity-qa}, LongVideoBench~\citep{longvideobench}, PerceptionTest~\citep{perpceptiontest}, 
NExT-QA~\citep{next-qa}, VNBench~\citep{vnbench}, and VideoMME~\citep{video-mme}. LongVILA-R1-7B outperforms LongVILA-7B across all these benchmarks, with varying margins.
}}}
\resizebox{\linewidth}{!}
{
\begin{tabular}{l|cccccccc}
\toprule
\multirow{2}{*}{Model} & {Act.Net-QA}   & {L.V.Bench} & {Per.Test}      & {NExT-QA}       & {VNBench}   & \multicolumn{2}{c}{VideoMME} \\ 
     & test                & val            & val            & mc            & val       & w/o sub.  & w/ sub.   \\ \midrule
GPT-4o mini                                                        &  -        &    56.5        &    -          &   -        &   -    &    64.8     &   68.9      \\
GPT-4o                                                        &  61.9         &   66.7         &     -           &    -       &    64.4   &     71.9    &   77.2      \\
Gemini-1.5-Pro                                                         &  57.5       &   64.0         &   -         &    -       &   66.7    &    75.0     &   81.3      \\ \midrule
Video-LLaVA-7B                                                                  & 45.3       & 37.6           & -              & -             & 12.4      & 39.9          & 41.6          \\
Flash-VStream-7B                                                                  & 51.9              & -              & -             & 61.6          & -         & -             & -             \\
ShareGPT4Video-8B                                                                 & 50.8                & 41.8           & -              & -             & -         & 39.9          & 43.6          \\
VideoLLaMA2-7B                                                                 & 50.2         & -              & 51.4           & -             & 4.5       & 47.9          & 50.3          \\
VideoLLaMA2.1-7B                                                                 & 53.0             & -              & 54.9           & -             & -         & 54.9          & 56.4          \\
Kangaroo-8B                                                                 & -         & 54.8           & -           & -             & -         & 56.0          & 57.6          \\
PLLaVA-7B                                                                & 56.3              & 39.2           & -           & -             & -         & -             & -             \\
LLaVA-OV-7B                                                                & 56.7              & 56.4           & 57.1             & 79.4          & 51.8      & 58.2          & 61.5          \\ \midrule
LongVILA-7B                                                                 & {59.5}  & {57.1}  & {58.1} & {80.7} & {63.0} & {60.1} & {65.1} \\ 
LongVILA-R1-7B                                                                  &  \textbf{64.8} & \textbf{58.0}  & \textbf{68.9} & \textbf{81.5} & \textbf{75.5} & \textbf{65.1} & \textbf{71.1} \\
\bottomrule
\end{tabular}}
\label{tab:9-benchmarks}
\end{center}
\end{table*}

\begin{table*}[t]
\begin{center}
\caption{{Performance on LongVideo-Reason-eval. LongVILA-R1-7B achieves a strong overall score.}}
\setlength{\tabcolsep}{5.5mm}{
\resizebox{\linewidth}{!}{
\begin{tabular}{l|cccc|c}
\toprule
Model          & Temporal & Goal & Plot & Spatial & Overall \\ \midrule
Video-R1-7B~\cite{feng2025video}    &  61.4  & 85.0  & 62.0  &  \textbf{58.5}   & 68.1    \\ 
Gemini-1.5-Pro~\cite{team2024gemini} & 65.4  &  81.9    & 67.8 &  53.3  & 69.3    \\ \midrule
LongVILA-7B & 58.0  &  80.2 & 57.1 & 46.7  & 62.7    \\
LongVILA-R1-7B &  \textbf{68.1}  &  \textbf{85.7}  & \textbf{70.6}  &  53.3  & \textbf{72.0}    \\ \bottomrule
\end{tabular}
}
}
\label{tab:LongVideo-Reason}
\end{center}
\end{table*}

\begin{table*}[t]
\begin{center}
\caption{{Performance comparison on VideoMME~\citep{video-mme} benchmark in details.}}
\resizebox{\linewidth}{!}{
\begin{tabular}{l|cccc|cccc}
\toprule
\multirow{2}{*}{Model}  & \multicolumn{4}{c}{\textit{w/o subtitle}} & \multicolumn{4}{c}{\textit{w subtitle}} \\
                        & Overall    & Short    & Medium    & Long   & Overall   & Short   & Medium   & Long   \\ \midrule
Video-R1-7B~\cite{feng2025video}                                             &  61.4  &   -   &   -   &  -  &   -   &  -   &   -   &   - \\
Apollo-7B~\cite{apollo}                                             &  61.1  &   -   &   -   &  -  &   63.3   &  -   &   -   &   - \\
LLaVA-Video-7B~\cite{llava-video}                                             &  63.3  &   -   &   -   &  -  &   69.7   &  -   &   -   &   - \\
NVILA-8B-Video~\cite{liu2024nvila}                                             &  64.2  &   -   &   -   &  -  &   70.0   &  -   &   -   &   - \\
Video-LLaVA-7B~\cite{lin2023video}                                                   & 39.9       & 45.3     & 38.0      & 36.2   & 41.6      & 46.1    & 40.7     & 38.1   \\
SliME-8B~\cite{zhang2024beyond}                                                    & 45.3       & 53.3     & 55.4      & 39.8   & 47.2      & 55.4    & 44.4     & 41.7   \\
ShareGPT4Video-8B~\cite{chen2024sharegpt4video}                                       & 39.9       & 48.3     & 36.3      & 35.0   & 43.6      & 53.6    & 39.3     & 37.9   \\
VideoChat2-7B~\cite{li2023videochat}                                         & 39.5       & 48.3     & 37.0      & 33.2   & 43.8      & 52.8    & 39.4     & 39.2   \\
Chat-Univi-v1.5-7B~\cite{jin2024chat}                                          & 40.6       & 45.7     & 40.3      & 35.8   & 45.9      & 51.2    & 44.6     & 41.8   \\
Kangaroo-8B~\cite{liu2024kangaroo}                                              & {56.0}       & 66.1     & 55.3      & 46.7   & {57.6}      & {68.0}    & 55.4     & {49.3}   \\ 
ShareGemini-7B~\cite{wang2024retake}                                       & 43.2       & 49.1     & 41.3      & 39.1   & 47.9      & 49.1    & 47.3     & 43.4   \\
LongVA-7B~\cite{zhang2024long}                                             & 52.6       & 61.1     & 50.4      & 46.2   & 54.3      & 61.1    & 53.6     & 47.6   \\
VITA-1.5-7B~\cite{vita-1.5}                                             & 56.1       & 67.0     & 54.2      & 47.1   & 58.7      & 69.9    & 55.7     & 50.4   \\
\midrule
  LongVILA-7B~\cite{chen2024longvila}      & {{60.1}}       & {{69.0}}     & {{58.3}}      & {{53.0}}   & {{65.1}}      & {{72.9}}    & {{64.9}}     & {{57.4}}   \\
LongVILA-R1-7B   &   \textbf{65.1}     &  \textbf{76.8}   &  \textbf{63.2}    &  \textbf{55.2}  &   \textbf{71.1}   &  \textbf{79.2}   &  \textbf{69.7}    &  \textbf{64.3}  \\ 
\bottomrule
\end{tabular}}
\label{tab:video-mme}
\end{center}
\end{table*}

\paragraph{Ablation on pipeline and datasets.} 
As shown in Table~\ref{tab:ablation-training}, we ablate the effectiveness of training stages and datasets, using LongVILA-1.5B as a starting point. The accuracies are evaluated on LongVideo-Reason-eval. \xmark $\,$ means skipping this stage, \cmark $\,$ means training this stage with our datasets, and \textbf{O} means training this stage with other datasets~\cite{llava-video,next-qa,perpceptiontest,clevr,star}. Our CoT-SFT dataset results in a better performance than that of other datasets. In addition, incorporating RL on top of the warm-up phase (CoT-SFT) yields additional improvements compared to using only SFT.
We show that if we skip CoT-SFT and train our models directly with RL, the accuracy drops. If we apply Video-R1 datasets in both CoT-SFT and RL stages, the performance is inferior to using ours.

\paragraph{Training efficiency on MR-SP.} 
We conduct the training efficiency comparison for our MR-SP system, one A100 node, {\em i.e.}, 8xA100 (80GB) GPUs. We measure the forward time for each training step. The results are obtained after 10 warming up iterations and averaged over 5 iterations. We use LongVILA-7B-R1 model with training batch size as 1 per GPU and rollout number as 5. 

Figure~\ref{fig:efficiency-mrsp} presents the training efficiency comparison across different numbers of frames. The figure plots the runtime per step (in seconds) for three settings: the plain RL system without MR-SP, MR-SP Stage 1 only, and the full MR-SP system (Stage 1 \& 2). The baseline runtime increases steeply as the frame count grows. Using only Stage 1 of MR-SP significantly improves efficiency up to 512 frames but encounters GPU out-of-memory (OOM) issues beyond that point. In contrast, the full MR-SP system consistently reduces runtime, achieving up to a 2.1× speedup at 512 frames and scaling efficiently to 1024 frames without OOM, highlighting the benefit of combining sequence reuse and sequence parallelism for long video RL training.

\begin{table}[t]
\begin{center}
\begin{minipage}{0.48\textwidth} 
\centering
\caption{{Ablations on frames and reasoning (R.).}}
\resizebox{\linewidth}{!}{
\begin{tabular}{l|cccccc}
\toprule
Frames & 16   & 32   & 64   & 128  & 256  & 512  \\ \midrule
w/o R. & 55.7 & 56.4 & 58.1 & 60.5 & 60.7 & 60.2 \\
w/ R.  & 55.9 & 56.7 & 61.9 & 62.6 & 64.1 & 64.3 \\ \bottomrule
\end{tabular}}
\label{tab:ablation-training}
\end{minipage}
\hfill
\begin{minipage}{0.5\textwidth} 
\centering
\caption{{Ablations on CoT-SFT and RL.}}
\resizebox{\linewidth}{!}{
\begin{tabular}{c|cccccc}
\toprule
CoT-SFT  &     \xmark                 & \cmark      & \xmark                  &              \textbf{O}           &  \textbf{O}           & \cmark                     \\
RL       & \xmark & \xmark   & \cmark    & \xmark  &  \textbf{O}           & \cmark \\ \midrule
Accuracy & 58.1                 & 60.2        & 52.4             &   59.1            &  59.4                 & 61.9                    \\ \bottomrule
\end{tabular}}
\label{tab:scaling-video-frames}
\end{minipage}
\end{center}
\end{table}

\section{Conclusion}
We present a comprehensive framework designed to fully scale VLMs for reasoning over long videos. LongVILA-R1 encompasses a meticulously constructed large-scale dataset, LongVideo-Reason, and a parallelized training framework, MR-SP. Leveraging our curated dataset of 104K long video question-reasoning-answer pairs, combined with other open-source video datasets, we adopt a two-stage training process that integrates CoT-SFT and RL. 
LongVILA-R1-7B demonstrates outstanding performance on mainstream video benchmarks, achieving 65.1\% and 71.1\% on VideoMME without and with subtitles.
LongVILA-R1 supports processing up to 8,192 video frames per video, and configurable FPS settings. 
Notably, our MR-SP leads to a speed up of 2.1$\times$ for long video RL training, supporting hour-level (3,600 frames) RL training on a single node of 8 A100 GPUs.
In addition, we release our training system to the public, which supports RL training across multiple modalities (video, text, and audio), various models (including VILA and Qwen series), and even image and video generation models.

\begin{figure*}[!t]
\centerline{\includegraphics[width=1\textwidth]{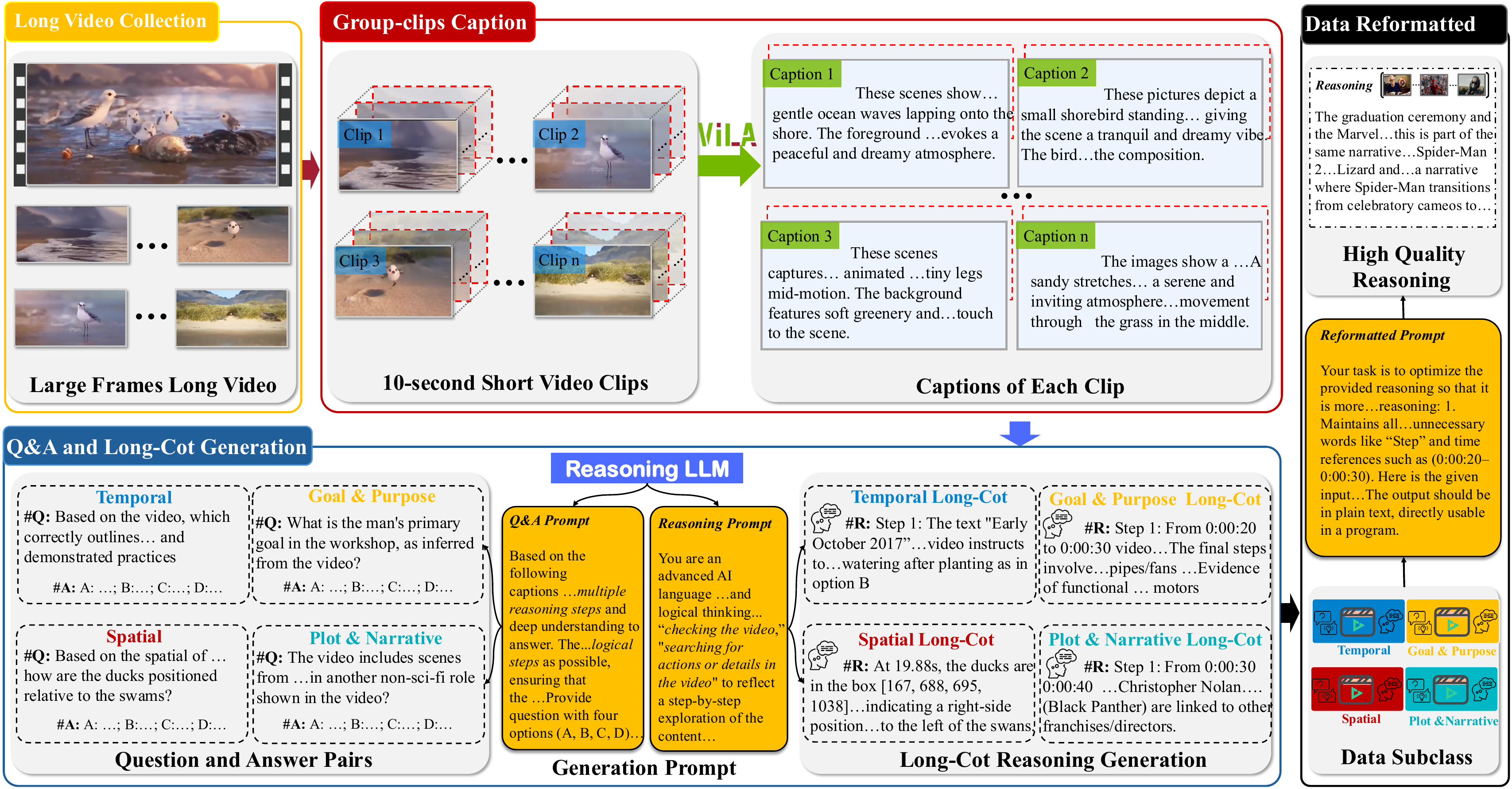}}
\caption{Detailed data generation process for the LongVideo-Reason dataset, supplementing Figure~\ref{fig:data-generation}.
}
\label{fig:data-generation-appendix}
\end{figure*}

\paragraph{Limitations}
While our RL training system is capable of handling long video frames efficiently (3,600 frames on a single node of A100 GPUs), scaling to significantly longer sequences, more modalities (like include audio for omni VLMs) or large batch sizes would require distributed training across multiple GPUs. This requires more GPUs, making it less feasible to run on limited resources.

\section{Border Impacts}
The development of LongVILA-R1 represents a significant advancement in long video reasoning, enabling real world systems to perform sophisticated temporal and compositional understanding across diverse and extended visual contexts. This progress holds transformative potential across multiple domains, including embodied AI, robotics, autonomous systems, and AR/VR applications. By equipping vision-language models (VLMs) with the ability to process and reason over long-duration video data, LongVILA-R1 lays the foundation for AI systems capable of understanding event sequences, and inferring causal and physical relationships over extended frames. 

Long video reasoning technologies powered by LongVILA-R1 can significantly enhance embodied AI and robotics by enabling agents to sustain coherent, long-term understanding of their environment. Robots equipped with such capabilities would excel in performing complex, multi-stage tasks, adapting to dynamic contexts, and building richer world models for planning and decision-making. These advancements also promise to unlock new opportunities in education, healthcare, and entertainment. For instance, long video understanding could enable AI tutors to analyze and summarize extended instructional videos, or assist healthcare professionals in reviewing lengthy procedural recordings. Furthermore, such systems could enhance sports analytics, and other areas requiring nuanced temporal reasoning.

Figure~\ref{fig:data-generation-appendix} shows the the data-generation where our supervision is built from \emph{grouped short clips} that form long videos, per-clip \emph{captions}, and LLM-generated \emph{Q\&A with Long-CoT} that is formatted as multiple-choice reasoning rather than identity labels or free-form attribution. This design choice intentionally steers the supervision signal toward temporal and causal evidence (what happens, when, and why) and away from sensitive attributes (who a person is), thereby reducing privacy and profiling risks at the data layer. Concretely, we derive questions from captions and temporally grounded observations instead of collecting personal metadata; we avoid tasks that require demographic inference, face recognition, or persistent identity tracking; and we release structured Q\&A rather than user-originating personal content. In this way, the dataset construction supports the intended societal benefits of long-video reasoning while mitigating risks such as privacy leakage, surveillance-like use, and stereotype amplification.

In conclusion, LongVILA-R1 demonstrates the potential of long video reasoning to drive progress across a wide range of applications, from robotics to immersive virtual environments. However, unlocking the full promise of this technology requires a steadfast commitment to ethical principles, privacy protection, and the broader goal of benefiting humanity. By addressing these challenges, the AI community can ensure that advancements in long video reasoning contribute positively to society while mitigating associated risks.

{
  \bibliography{neurips_2025}

\begin{thebibliography}{10}

\bibitem{chen2025r1v}
Liang Chen, Lei Li, Haozhe Zhao, Yifan Song, and Vinci.
\newblock R1-v: Reinforcing super generalization ability in vision-language models with less than \$3, 2025.
\newblock Accessed: 2025-02-02.

\bibitem{chen2024sharegpt4video}
Lin Chen, Xilin Wei, Jinsong Li, Xiaoyi Dong, Pan Zhang, Yuhang Zang, Zehui Chen, Haodong Duan, Zhenyu Tang, Li~Yuan, et~al.
\newblock Sharegpt4video: Improving video understanding and generation with better captions.
\newblock {\em NeurIPS}, 37:19472--19495, 2024.

\bibitem{InternEvo}
Qiaoling Chen, Diandian Gu, Guoteng Wang, Xun Chen, YingTong Xiong, Ting Huang, Qinghao Hu, Xin Jin, Yonggang Wen, Tianwei Zhang, and Peng Sun.
\newblock Internevo: Efficient long-sequence large language model training via hybrid parallelism and redundant sharding.
\newblock {\em CoRR}, abs/2401.09149, 2024.

\bibitem{longlora}
Yukang Chen, Shengju Qian, Haotian Tang, Xin Lai, Zhijian Liu, Song Han, and Jiaya Jia.
\newblock Longlora: Efficient fine-tuning of long-context large language models.
\newblock In {\em ICLR}, 2024.

\bibitem{internvl2.5}
Zhe Chen, Weiyun Wang, Yue Cao, Yangzhou Liu, Zhangwei Gao, Erfei Cui, Jinguo Zhu, Shenglong Ye, Hao Tian, Zhaoyang Liu, Lixin Gu, Xuehui Wang, Qingyun Li, Yimin Ren, Zixuan Chen, Jiapeng Luo, Jiahao Wang, Tan Jiang, Bo~Wang, Conghui He, Botian Shi, Xingcheng Zhang, Han Lv, Yi~Wang, Wenqi Shao, Pei Chu, Zhongying Tu, Tong He, Zhiyong Wu, Huipeng Deng, Jiaye Ge, Kai Chen, Min Dou, Lewei Lu, Xizhou Zhu, Tong Lu, Dahua Lin, Yu~Qiao, Jifeng Dai, and Wenhai Wang.
\newblock Expanding performance boundaries of open-source multimodal models with model, data, and test-time scaling.
\newblock {\em CoRR}, abs/2412.05271, 2024.

\bibitem{guo2025deepseek}
DeepSeek{-}AI.
\newblock Deepseek-r1: Incentivizing reasoning capability in llms via reinforcement learning.
\newblock {\em CoRR}, abs/2501.12948, 2025.

\bibitem{USP}
Jiarui Fang and Shangchun Zhao.
\newblock {USP:} {A} unified sequence parallelism approach for long context generative {AI}.
\newblock {\em CoRR}, abs/2405.07719, 2024.

\bibitem{feng2025video}
Kaituo Feng, Kaixiong Gong, Bohao Li, Zonghao Guo, Yibing Wang, Tianshuo Peng, Benyou Wang, and Xiangyu Yue.
\newblock Video-r1: Reinforcing video reasoning in mllms.
\newblock {\em CoRR}, abs/2503.21776, 2025.

\bibitem{video-mme}
Chaoyou Fu, Yuhan Dai, Yondong Luo, Lei Li, Shuhuai Ren, Renrui Zhang, Zihan Wang, Chenyu Zhou, Yunhang Shen, Mengdan Zhang, Peixian Chen, Yanwei Li, Shaohui Lin, Sirui Zhao, Ke~Li, Tong Xu, Xiawu Zheng, Enhong Chen, Rongrong Ji, and Xing Sun.
\newblock Video-mme: The first-ever comprehensive evaluation benchmark of multi-modal llms in video analysis.
\newblock {\em CoRR}, abs/2405.21075, 2024.

\bibitem{vita-1.5}
Chaoyou Fu, Haojia Lin, Xiong Wang, Yifan Zhang, Yunhang Shen, Xiaoyu Liu, Haoyu Cao, Zuwei Long, Heting Gao, Ke~Li, Long Ma, Xiawu Zheng, Rongrong Ji, Xing Sun, Caifeng Shan, and Ran He.
\newblock {VITA-1.5:} towards gpt-4o level real-time vision and speech interaction.
\newblock {\em CoRR}, abs/2501.01957, 2025.

\bibitem{LoongTrain}
Diandian Gu, Peng Sun, Qinghao Hu, Ting Huang, Xun Chen, Yingtong Xiong, Guoteng Wang, Qiaoling Chen, Shangchun Zhao, Jiarui Fang, Yonggang Wen, Tianwei Zhang, Xin Jin, and Xuanzhe Liu.
\newblock Loongtrain: Efficient training of long-sequence llms with head-context parallelism.
\newblock {\em CoRR}, pdf/2406.18485, 2024.

\bibitem{guo2025fila}
Yanan Guo, Wenhui Dong, Jun Song, Shiding Zhu, Xuan Zhang, Hanqing Yang, Yingbo Wang, Yang Du, Xianing Chen, and Bo~Zheng.
\newblock Fila-video: Spatio-temporal compression for fine-grained long video understanding.
\newblock {\em CoRR}, abs/2504.20384, 2025.

\bibitem{han2023shot2story20k}
Mingfei Han, Linjie Yang, Xiaojun Chang, and Heng Wang.
\newblock Shot2story20k: {A} new benchmark for comprehensive understanding of multi-shot videos.
\newblock {\em CoRR}, abs/2312.10300, 2023.

\bibitem{han2024videoespresso}
Songhao Han, Wei Huang, Hairong Shi, Le~Zhuo, Xiu Su, Shifeng Zhang, Xu~Zhou, Xiaojuan Qi, Yue Liao, and Si~Liu.
\newblock Videoespresso: {A} large-scale chain-of-thought dataset for fine-grained video reasoning via core frame selection.
\newblock {\em CoRR}, abs/2411.14794, 2024.

\bibitem{huang2025vision}
Wenxuan Huang, Bohan Jia, Zijie Zhai, Shaosheng Cao, Zheyu Ye, Fei Zhao, Zhe Xu, Yao Hu, and Shaohui Lin.
\newblock Vision-r1: Incentivizing reasoning capability in multimodal large language models.
\newblock {\em CoRR}, abs/2503.06749, 2025.

\bibitem{jacobs2023deepspeed}
Sam~Ade Jacobs, Masahiro Tanaka, Chengming Zhang, Minjia Zhang, Shuaiwen~Leon Song, Samyam Rajbhandari, and Yuxiong He.
\newblock Deepspeed ulysses: System optimizations for enabling training of extreme long sequence transformer models.
\newblock {\em CoRR}, abs/2309.14509, 2023.

\bibitem{jin2024chat}
Peng Jin, Ryuichi Takanobu, Wancai Zhang, Xiaochun Cao, and Li~Yuan.
\newblock Chat-univi: Unified visual representation empowers large language models with image and video understanding.
\newblock In {\em CVPR}, pages 13700--13710, 2024.

\bibitem{clevr}
Justin Johnson, Bharath Hariharan, Laurens van~der Maaten, Li~Fei{-}Fei, C.~Lawrence Zitnick, and Ross~B. Girshick.
\newblock {CLEVR:} {A} diagnostic dataset for compositional language and elementary visual reasoning.
\newblock In {\em CVPR}, pages 1988--1997, 2017.

\bibitem{kwon2023efficient}
Woosuk Kwon, Zhuohan Li, Siyuan Zhuang, Ying Sheng, Lianmin Zheng, Cody~Hao Yu, Joseph~E. Gonzalez, Hao Zhang, and Ion Stoica.
\newblock Efficient memory management for large language model serving with pagedattention.
\newblock In {\em Proceedings of the ACM SIGOPS 29th Symposium on Operating Systems Principles}, 2023.

\bibitem{li2023lightseq}
Dacheng Li, Rulin Shao, Anze Xie, Eric Xing, Joseph~E Gonzalez, Ion Stoica, Xuezhe Ma, and Hao Zhang.
\newblock Lightseq: Sequence level parallelism for distributed training of long context transformers.
\newblock 2023.

\bibitem{li2023videochat}
Kunchang Li, Yinan He, Yi~Wang, Yizhuo Li, Wenhai Wang, Ping Luo, Yali Wang, Limin Wang, and Yu~Qiao.
\newblock Videochat: Chat-centric video understanding.
\newblock {\em CoRR}, abs/2305.06355, 2023.

\bibitem{lin2023video}
Bin Lin, Yang Ye, Bin Zhu, Jiaxi Cui, Munan Ning, Peng Jin, and Li~Yuan.
\newblock Video-llava: Learning united visual representation by alignment before projection.
\newblock In {\em EMNLP}, pages 5971--5984. ACL, 2024.

\bibitem{lin2024vila}
Ji~Lin, Hongxu Yin, Wei Ping, Pavlo Molchanov, Mohammad Shoeybi, and Song Han.
\newblock Vila: On pre-training for visual language models.
\newblock In {\em CVPR}, pages 26689--26699, 2024.

\bibitem{liu2024codemind}
Changshu Liu, Shizhuo~Dylan Zhang, and Reyhaneh Jabbarvand.
\newblock Codemind: {A} framework to challenge large language models for code reasoning.
\newblock {\em CoRR}, abs/2402.09664, 2024.

\bibitem{liu2023ring}
Hao Liu, Matei Zaharia, and Pieter Abbeel.
\newblock Ring attention with blockwise transformers for near-infinite context.
\newblock {\em CoRR}, abs/2310.01889, 2023.

\bibitem{liu2024kangaroo}
Jiajun Liu, Yibing Wang, Hanghang Ma, Xiaoping Wu, Xiaoqi Ma, Xiaoming Wei, Jianbin Jiao, Enhua Wu, and Jie Hu.
\newblock Kangaroo: {A} powerful video-language model supporting long-context video input.
\newblock {\em CoRR}, abs/2408.15542, 2024.

\bibitem{liu2024nvila}
Zhijian Liu, Ligeng Zhu, Baifeng Shi, Zhuoyang Zhang, Yuming Lou, Shang Yang, Haocheng Xi, Shiyi Cao, Yuxian Gu, Dacheng Li, et~al.
\newblock {NVILA:} efficient frontier visual language models.
\newblock {\em CoRR}, abs/2412.04468, 2024.

\bibitem{moritz2018ray}
Philipp Moritz, Robert Nishihara, Stephanie Wang, Alexey Tumanov, Richard Liaw, Eric Liang, Melih Elibol, Zongheng Yang, William Paul, Michael~I Jordan, et~al.
\newblock Ray: A distributed framework for emerging $\{$AI$\}$ applications.
\newblock In {\em OSDI}, pages 561--577, 2018.

\bibitem{gpt4o}
OpenAI.
\newblock Gpt-4o.
\newblock 2025.

\bibitem{ouyang2022training}
Long Ouyang, Jeffrey Wu, Xu~Jiang, Diogo Almeida, Carroll Wainwright, Pamela Mishkin, Chong Zhang, Sandhini Agarwal, Katarina Slama, Alex Ray, et~al.
\newblock Training language models to follow instructions with human feedback.
\newblock {\em NeurIPS}, 35:27730--27744, 2022.

\bibitem{perpceptiontest}
Viorica Patraucean, Lucas Smaira, Ankush Gupta, Adri{\`{a}} Recasens, Larisa Markeeva, Dylan Banarse, Skanda Koppula, Joseph Heyward, Mateusz Malinowski, Yi~Yang, Carl Doersch, Tatiana Matejovicova, Yury Sulsky, Antoine Miech, Alexandre Fr{\'{e}}chette, Hanna Klimczak, Raphael Koster, Junlin Zhang, Stephanie Winkler, Yusuf Aytar, Simon Osindero, Dima Damen, Andrew Zisserman, and Jo{\~{a}}o Carreira.
\newblock Perception test: {A} diagnostic benchmark for multimodal video models.
\newblock In {\em NeurIPS}, 2023.

\bibitem{peng2025lmm}
Yingzhe Peng, Gongrui Zhang, Miaosen Zhang, Zhiyuan You, Jie Liu, Qipeng Zhu, Kai Yang, Xingzhong Xu, Xin Geng, and Xu~Yang.
\newblock {LMM-R1:} empowering 3b lmms with strong reasoning abilities through two-stage rule-based {RL}.
\newblock {\em CoRR}, abs/2503.07536, 2025.

\bibitem{dpo}
Rafael Rafailov, Archit Sharma, Eric Mitchell, Christopher~D. Manning, Stefano Ermon, and Chelsea Finn.
\newblock Direct preference optimization: Your language model is secretly a reward model.
\newblock In {\em NeurIPS}, 2023.

\bibitem{team2024gemini}
Machel Reid, Nikolay Savinov, Denis Teplyashin, Dmitry Lepikhin, Timothy~P. Lillicrap, Jean{-}Baptiste Alayrac, Radu Soricut, Angeliki Lazaridou, Orhan Firat, Julian Schrittwieser, Ioannis Antonoglou, Rohan Anil, Sebastian Borgeaud, Andrew~M. Dai, Katie Millican, Ethan Dyer, et~al.
\newblock Gemini 1.5: Unlocking multimodal understanding across millions of tokens of context.
\newblock {\em CoRR}, abs/2403.05530, 2024.

\bibitem{shao2024deepseekmath}
Zhihong Shao, Peiyi Wang, Qihao Zhu, Runxin Xu, Junxiao Song, Mingchuan Zhang, Y.~K. Li, Y.~Wu, and Daya Guo.
\newblock Deepseekmath: Pushing the limits of mathematical reasoning in open language models.
\newblock {\em CoRR}, abs/2402.03300, 2024.

\bibitem{grpo}
Zhihong Shao, Peiyi Wang, Qihao Zhu, Runxin Xu, Junxiao Song, Mingchuan Zhang, Y.~K. Li, Y.~Wu, and Daya Guo.
\newblock Deepseekmath: Pushing the limits of mathematical reasoning in open language models.
\newblock {\em CoRR}, abs/2402.03300, 2024.

\bibitem{shen2025long}
Yunhang Shen, Chaoyou Fu, Shaoqi Dong, Xiong Wang, Peixian Chen, Mengdan Zhang, Haoyu Cao, Ke~Li, Xiawu Zheng, Yan Zhang, et~al.
\newblock Long-vita: Scaling large multi-modal models to 1 million tokens with leading short-context accuray.
\newblock {\em CoRR}, abs/2502.05177, 2025.

\bibitem{sheng2024hybridflow}
Guangming Sheng, Chi Zhang, Zilingfeng Ye, Xibin Wu, Wang Zhang, Ru~Zhang, Yanghua Peng, Haibin Lin, and Chuan Wu.
\newblock Hybridflow: {A} flexible and efficient {RLHF} framework.
\newblock In {\em EuroSys}, pages 1279--1297. {ACM}, 2025.

\bibitem{vila-hd}
Baifeng Shi, Boyi Li, Han Cai, Yao Lu, Sifei Liu, Marco Pavone, Jan Kautz, Song Han, Trevor Darrell, Pavlo Molchanov, and Hongxu Yin.
\newblock Scaling vision pre-training to 4k resolution.
\newblock {\em CoRR}, abs/2503.19903, 2025.

\bibitem{team2025kimi}
Kimi Team, Angang Du, Bofei Gao, Bowei Xing, Changjiu Jiang, Cheng Chen, Cheng Li, Chenjun Xiao, Chenzhuang Du, Chonghua Liao, et~al.
\newblock Kimi k1.5: Scaling reinforcement learning with llms.
\newblock {\em CoRR}, abs/2501.12599, 2025.

\bibitem{qwen2-vl}
Peng Wang, Shuai Bai, Sinan Tan, Shijie Wang, Zhihao Fan, Jinze Bai, Keqin Chen, Xuejing Liu, Jialin Wang, Wenbin Ge, Yang Fan, Kai Dang, Mengfei Du, Xuancheng Ren, Rui Men, Dayiheng Liu, Chang Zhou, Jingren Zhou, and Junyang Lin.
\newblock Qwen2-vl: Enhancing vision-language model's perception of the world at any resolution.
\newblock {\em CoRR}, abs/2409.12191, 2024.

\bibitem{wang2024retake}
Xiao Wang, Qingyi Si, Jianlong Wu, Shiyu Zhu, Li~Cao, and Liqiang Nie.
\newblock Retake: Reducing temporal and knowledge redundancy for long video understanding.
\newblock {\em CoRR}, abs/2412.20504, 2024.

\bibitem{weng2024longvlm}
Yuetian Weng, Mingfei Han, Haoyu He, Xiaojun Chang, and Bohan Zhuang.
\newblock Longvlm: Efficient long video understanding via large language models.
\newblock In {\em ECCV}, pages 453--470. Springer, 2024.

\bibitem{star}
Bo~Wu, Shoubin Yu, Zhenfang Chen, Josh Tenenbaum, and Chuang Gan.
\newblock {STAR:} {A} benchmark for situated reasoning in real-world videos.
\newblock In {\em NeurIPS - Datasets and Benchmarks}, 2021.

\bibitem{longvideobench}
Haoning Wu, Dongxu Li, Bei Chen, and Junnan Li.
\newblock Longvideobench: {A} benchmark for long-context interleaved video-language understanding.
\newblock {\em CoRR}, abs/2407.15754, 2024.

\bibitem{next-qa}
Junbin Xiao, Xindi Shang, Angela Yao, and Tat{-}Seng Chua.
\newblock Next-qa: Next phase of question-answering to explaining temporal actions.
\newblock In {\em CVPR}, pages 9777--9786, 2021.

\bibitem{chen2024longvila}
Fuzhao Xue, Yukang Chen, Dacheng Li, Qinghao Hu, Ligeng Zhu, Xiuyu Li, Yunhao Fang, Haotian Tang, Shang Yang, Zhijian Liu, Ethan He, Hongxu Yin, Pavlo Molchanov, Jan Kautz, Linxi Fan, Yuke Zhu, Yao Lu, and Song Han.
\newblock Longvila: Scaling long-context visual language models for long videos.
\newblock In {\em ICLR}, 2025.

\bibitem{yang2024qwen2}
An~Yang, Baosong Yang, Beichen Zhang, Binyuan Hui, Bo~Zheng, Bowen Yu, Chengyuan Li, Dayiheng Liu, Fei Huang, Haoran Wei, et~al.
\newblock Qwen2.5 technical report.
\newblock {\em CoRR}, abs/2412.15115, 2024.

\bibitem{activity-qa}
Zhou Yu, Dejing Xu, Jun Yu, Ting Yu, Zhou Zhao, Yueting Zhuang, and Dacheng Tao.
\newblock Activitynet-qa: {A} dataset for understanding complex web videos via question answering.
\newblock In {\em AAAI}, pages 9127--9134, 2019.

\bibitem{videollama3}
Boqiang Zhang, Kehan Li, Zesen Cheng, Zhiqiang Hu, Yuqian Yuan, Guanzheng Chen, Sicong Leng, Yuming Jiang, Hang Zhang, Xin Li, Peng Jin, Wenqi Zhang, Fan Wang, Lidong Bing, and Deli Zhao.
\newblock Videollama 3: Frontier multimodal foundation models for image and video understanding.
\newblock {\em CoRR}, abs/2501.13106, 2025.

\bibitem{zhang2024long}
Peiyuan Zhang, Kaichen Zhang, Bo~Li, Guangtao Zeng, Jingkang Yang, Yuanhan Zhang, Ziyue Wang, Haoran Tan, Chunyuan Li, and Ziwei Liu.
\newblock Long context transfer from language to vision.
\newblock {\em CoRR}, abs/2406.16852, 2024.

\bibitem{zhang2024beyond}
Yifan Zhang, Qingsong Wen, Chaoyou Fu, Xue Wang, Zhang Zhang, Liang Wang, and Rong Jin.
\newblock Beyond llava-hd: Diving into high-resolution large multimodal models.
\newblock {\em CoRR}, abs/2406.08487, 2024.

\bibitem{llava-video}
Yuanhan Zhang, Jinming Wu, Wei Li, Bo~Li, Zejun Ma, Ziwei Liu, and Chunyuan Li.
\newblock Video instruction tuning with synthetic data.
\newblock {\em CoRR}, abs/2410.02713, 2024.

\bibitem{vnbench}
Zijia Zhao, Haoyu Lu, Yuqi Huo, Yifan Du, Tongtian Yue, Longteng Guo, Bingning Wang, Weipeng Chen, and Jing Liu.
\newblock Needle in {A} video haystack: {A} scalable synthetic framework for benchmarking video mllms.
\newblock {\em CoRR}, abs/2406.09367, 2024.

\bibitem{EasyR1}
Yaowei Zheng, Junting Lu, Shenzhi Wang, Zhangchi Feng, Dongdong Kuang, and Yuwen Xiong.
\newblock Easyr1: An efficient, scalable, multi-modality rl training framework, 2025.

\bibitem{zhu2025internvl3}
Jinguo Zhu, Weiyun Wang, Zhe Chen, Zhaoyang Liu, Shenglong Ye, Lixin Gu, Hao Tian, Yuchen Duan, Weijie Su, Jie Shao, Zhangwei Gao, Erfei Cui, Xuehui Wang, et~al.
\newblock Internvl3: Exploring advanced training and test-time recipes for open-source multimodal models.
\newblock {\em CoRR}, abs/2504.10479, 2025.

\bibitem{apollo}
Orr Zohar, Xiaohan Wang, Yann Dubois, Nikhil Mehta, Tong Xiao, Philippe Hansen{-}Estruch, Licheng Yu, Xiaofang Wang, Felix Juefei{-}Xu, Ning Zhang, Serena Yeung{-}Levy, and Xide Xia.
\newblock Apollo: An exploration of video understanding in large multimodal models.
\newblock In {\em CVPR}, pages 18891--18901, 2025.

\end{thebibliography}
  \bibliographystyle{plain}
}

\clearpage

\section*{NeurIPS Paper Checklist}
\begin{enumerate}

\item {\bf Claims}
    \item[] Question: Do the main claims made in the abstract and introduction accurately reflect the paper's contributions and scope?
    \item[] Answer: \answerYes{} 
    \item[] Justification: The main claims made in the abstract and introduction accurately reflect the paper's contributions and scope, as they align with the theoretical and experimental results presented in the paper and provide a clear understanding of the paper's goals.

\item {\bf Limitations}
    \item[] Question: Does the paper discuss the limitations of the work performed by the authors?
    \item[] Answer: \answerYes{} 
    \item[] Justification: The paper acknowledges the limitations of the work performed by the authors.

\item {\bf Theory assumptions and proofs}
    \item[] Question: For each theoretical result, does the paper provide the full set of assumptions and a complete (and correct) proof?
    \item[] Answer: \answerNA{} 
    \item[] Justification:

    \item {\bf Experimental result reproducibility}
    \item[] Question: Does the paper fully disclose all the information needed to reproduce the main experimental results of the paper to the extent that it affects the main claims and/or conclusions of the paper (regardless of whether the code and data are provided or not)?
    \item[] Answer: \answerYes{} 
    \item[] Justification: The paper fully discloses all the information needed to reproduce the main experimental results, ensuring that the main claims and conclusions can be independently verified. This includes providing relevant details, methodologies, and any necessary parameters or configurations for conducting the experiments.

\item {\bf Open access to data and code}
    \item[] Question: Does the paper provide open access to the data and code, with sufficient instructions to faithfully reproduce the main experimental results, as described in supplemental material?
    \item[] Answer: \answerYes{} 
    \item[] Justification: The paper will provide open access to the data and code necessary to reproduce the main experimental results. It also includes sufficient instructions in the supplemental material on how to faithfully replicate the experiments conducted in the paper.

\item {\bf Experimental setting/details}
    \item[] Question: Does the paper specify all the training and test details (e.g., data splits, hyperparameters, how they were chosen, type of optimizer, etc.) necessary to understand the results?
    \item[] Answer: \answerYes{} 
    \item[] Justification: The paper specifies all the training and test details necessary to understand the results. This includes information on data splits, hyperparameters, the methodology for selecting hyperparameters, the type of optimizer used, and any other relevant details that are crucial for replicating and comprehending the reported results.

\item {\bf Experiment statistical significance}
    \item[] Question: Does the paper report error bars suitably and correctly defined or other appropriate information about the statistical significance of the experiments?
    \item[] Answer: \answerYes{} 
    \item[] Justification: The paper provides suitable information about the statistical significance of the experiments. This indicates that the authors have appropriately addressed the need for statistical analysis and have reported the relevant measures to support the reliability and significance of their experimental findings.

\item {\bf Experiments compute resources}
    \item[] Question: For each experiment, does the paper provide sufficient information on the computer resources (type of compute workers, memory, time of execution) needed to reproduce the experiments?
    \item[] Answer: \answerYes{} 
    \item[] Justification: The paper provides sufficient information on the computer resources required to reproduce each experiment. This includes details such as the type of compute workers used, the amount of memory required, and the time taken for the execution of the experiments. This information allows for accurate replication of the experiments and provides transparency regarding the computational requirements of the study.
    
\item {\bf Code of ethics}
    \item[] Question: Does the research conducted in the paper conform, in every respect, with the NeurIPS Code of Ethics \url{https://neurips.cc/public/EthicsGuidelines}?
    \item[] Answer: \answerYes{} 
    \item[] Justification: The research conducted in the paper conforms, in every respect, with the NeurIPS Code of Ethics.

\item {\bf Broader impacts}
    \item[] Question: Does the paper discuss both potential positive societal impacts and negative societal impacts of the work performed?
    \item[] Answer: \answerYes{} 
    \item[] Justification: The paper discusses both potential positive and negative societal impacts of the work performed, especially in the resource limited application.
    
\item {\bf Safeguards}
    \item[] Question: Does the paper describe safeguards that have been put in place for responsible release of data or models that have a high risk for misuse (e.g., pretrained language models, image generators, or scraped datasets)?
    \item[] Answer: \answerNA{} 
    \item[] Justification: The paper poses no such risks

\item {\bf Licenses for existing assets}
    \item[] Question: Are the creators or original owners of assets (e.g., code, data, models), used in the paper, properly credited and are the license and terms of use explicitly mentioned and properly respected?
    \item[] Answer: \answerYes{} 
    \item[] Justification: The paper only use the opensource datasets with legal license.

\item {\bf New assets}
    \item[] Question: Are new assets introduced in the paper well documented and is the documentation provided alongside the assets?
    \item[] Answer: \answerYes{} 
    \item[] Justification: The new assets introduced in the paper are well documented and are the documentation provided alongside the assets.

\item {\bf Crowdsourcing and research with human subjects}
    \item[] Question: For crowdsourcing experiments and research with human subjects, does the paper include the full text of instructions given to participants and screenshots, if applicable, as well as details about compensation (if any)? 
    \item[] Answer: \answerNA{} 
    \item[] Justification: The paper does not involve crowdsourcing nor research with human subjects.

\item {\bf Institutional review board (IRB) approvals or equivalent for research with human subjects}
    \item[] Question: Does the paper describe potential risks incurred by study participants, whether such risks were disclosed to the subjects, and whether Institutional Review Board (IRB) approvals (or an equivalent approval/review based on the requirements of your country or institution) were obtained?
    \item[] Answer: \answerNA{} 
    \item[] Justification: The paper does not involve crowdsourcing nor research with human subjects.

\item {\bf Declaration of LLM usage}
    \item[] Question: Does the paper describe the usage of LLMs if it is an important, original, or non-standard component of the core methods in this research? Note that if the LLM is used only for writing, editing, or formatting purposes and does not impact the core methodology, scientific rigorousness, or originality of the research, declaration is not required.
    \item[] Answer: \answerNA{} 
    \item[] Justification: The core method development in this research does not involve LLMs as any important, original, or non-standard components.

\end{enumerate}

\newpage
\appendix

\section*{Appendix}

\begin{figure*}[t]
\centerline{\includegraphics[width=0.9\textwidth]{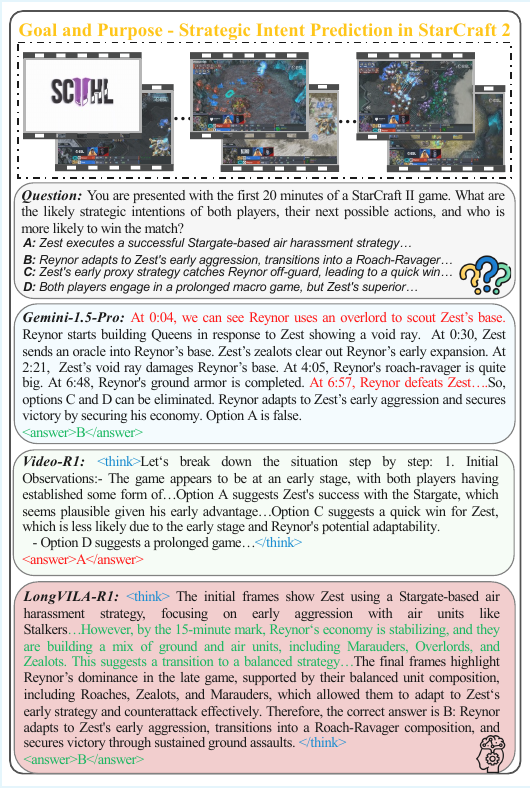}}
\caption{LongVILA-R1 reasoning example in the "Goal and Purpose" category. The video is available at \href{https://drive.google.com/file/d/1rnF4I6-EBpqhzA0SnwyajpxbAhMezDCn/view?usp=share_link}{Link}.}
\label{fig:demo_goal}
\end{figure*}

\begin{figure*}[t]
\centerline{\includegraphics[width=0.9\textwidth]{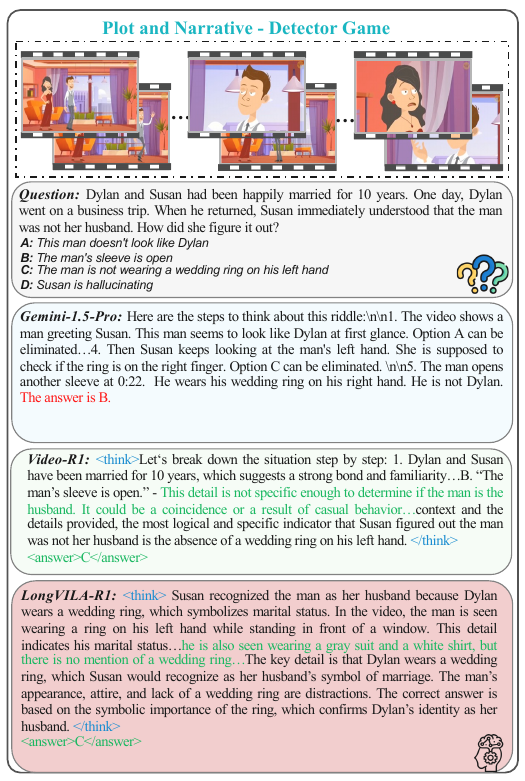}}
\caption{LongVILA-R1 reasoning example in the "Plot and Narrative" category. The video is available at \href{https://drive.google.com/file/d/1hCqYZd9uSjuAfuOk2zO5lJ4DosdKKLFj/view?usp=share_link}{Link}.}
\label{fig:demo_plot}
\end{figure*}

In this section, we selected examples of four types of reasoning tasks from the benchmark to evaluate and compare the reasoning processes and answers provided by Gemini-1.5-Pro, Video-R1-7B, and LongVILA-R1-7B. On Figure~\ref{fig:demo_goal}, a 20-minute StarCraft match is depicted, where the models analyze the players' unit compositions, strategies, and play styles to predict the potential developments on the battlefield. While Gemini-1.5-Pro produced a correct prediction of the outcome, its reasoning process contained factual inaccuracies. In contrast, Video-R1-7B, influenced by the characteristics of its training data, tended to summarize answers based on options, neglecting critical video details and resulting in incorrect reasoning. LongVILA-R1-7B, however, is able to accurately analyze the players' operational styles and specific moments marked in the video, leading to a comprehensive and accurate prediction of the match's trajectory. On Figure~\ref{fig:demo_plot}, another example demonstrates the models’ abilities in narrative reasoning and visual information analysis. Gemini-1.5-Pro failed to correctly infer why the man appearing for the second time in the video is not the husband. In contrast, both Video-R1-7B and LongVILA-R1-7B successfully reasoned that the man's habit of wearing a ring on his left hand is a key indicator, providing accurate answers.

\begin{figure*}[t]
\centerline{\includegraphics[width=0.9\textwidth]{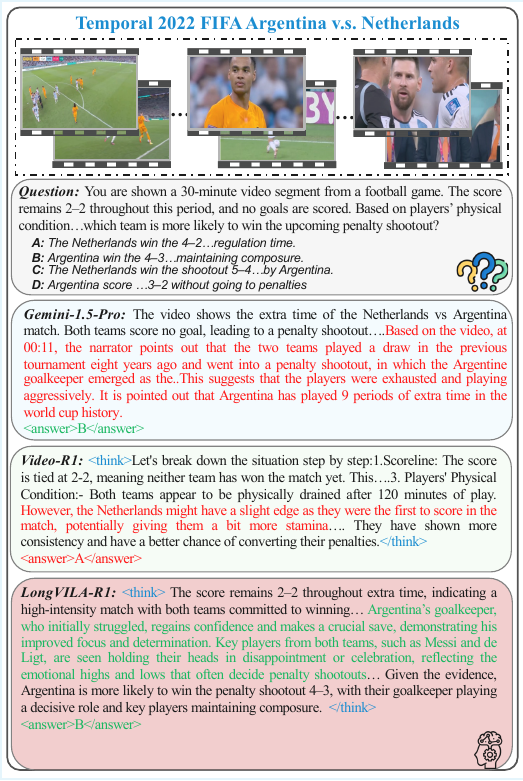}}
\caption{Detailed comparisons in the football game example. The video is available at \href{https://drive.google.com/file/d/1QJ-ZsDrmYS8v1XU4eWfYu5oHuXeyGSdK/view?usp=share_link}{Link}.}
\label{fig:demo_football}
\end{figure*}

\begin{figure*}[t]
\centerline{\includegraphics[width=0.9\textwidth]{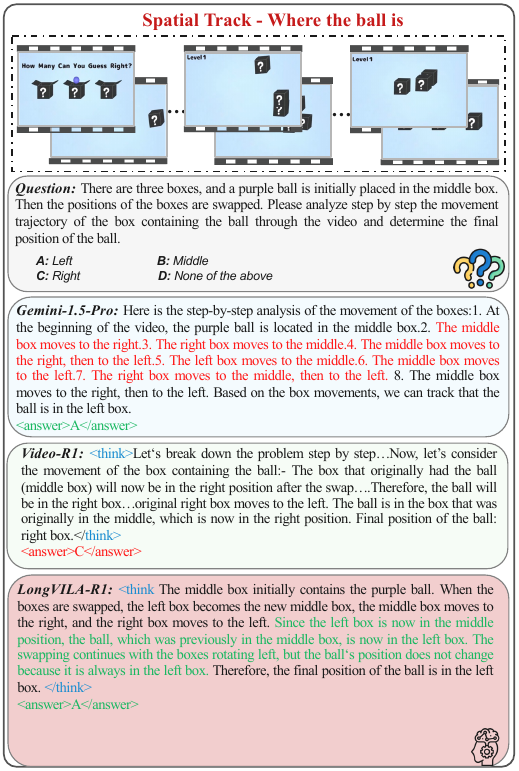}}
\caption{Detailed comparisons in the moving cup example. The video is available at \href{https://drive.google.com/file/d/1lo1E_bXXnMmWnFRudaSUgxMNxetEDHP9/view?usp=share_link}{Link}.}
\label{fig:demo_moving_cup}
\end{figure*}

\begin{figure*}[t]
\centerline{\includegraphics[width=0.9\textwidth]{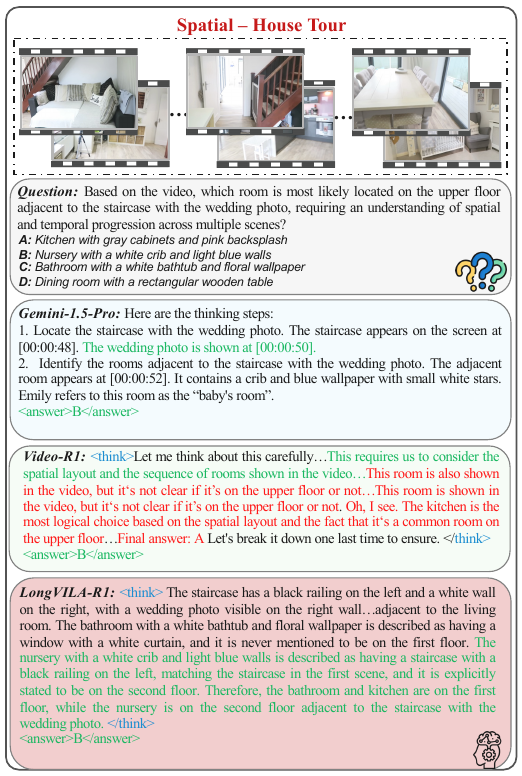}}
\caption{LongVILA-R1 reasoning example in the "Spatial" category. The video is available at \href{https://www.youtube.com/watch?v=ZG-Oef6Q2fk}{Link}.}
\label{fig:demo_spatial}
\end{figure*}

\begin{figure*}[t]
\centerline{\includegraphics[width=0.9\textwidth]{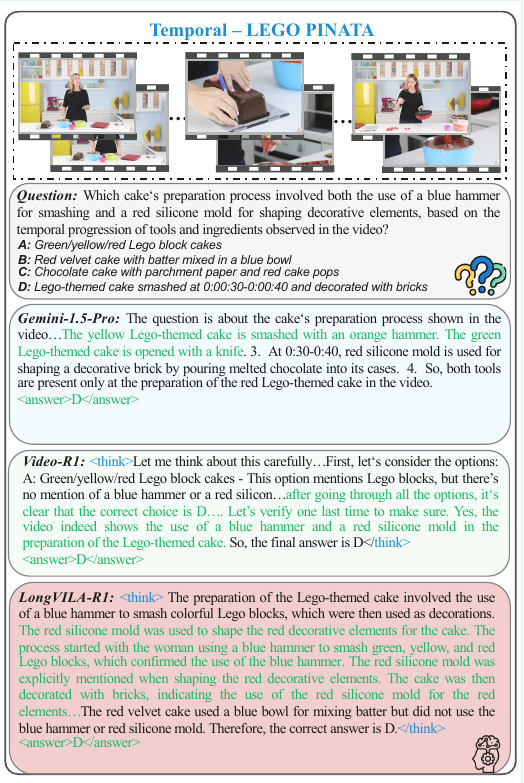}}
\caption{LongVILA-R1 reasoning example in the "Temporal" category. The video is available at \href{https://www.youtube.com/watch?v=co81n3wdN_4}{Link}.}
\label{fig:demo_temporal}
\end{figure*}

\begin{figure*}[t]
\centerline{\includegraphics[width=0.9\textwidth]{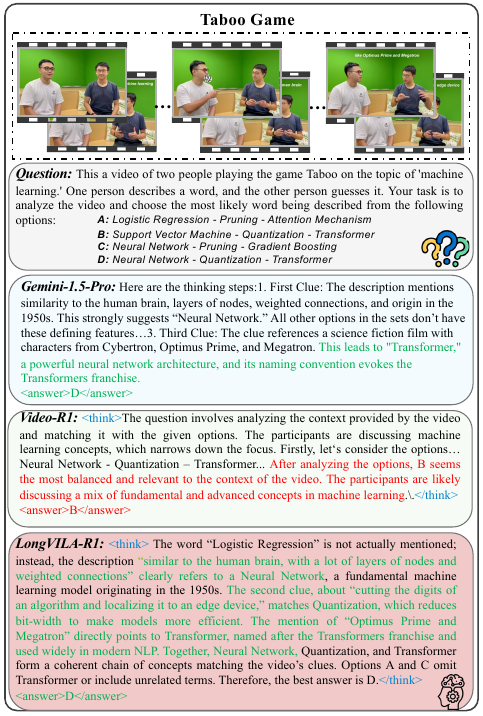}}
\caption{LongVILA-R1 reasoning example in the taboo game. The video is available at \href{https://drive.google.com/file/d/1jfvgVDNHqkGEBp3yd09HCQwBdd7K2QYs/view?usp=share_link}{Link}.}
\label{fig:demo_taboo}
\end{figure*}

Figure~\ref{fig:demo_spatial} illustrates the models' spatial perception and reasoning abilities as the camera moves through a room. Gemini-1.5-Pro effectively identified the key information within the video and provided the correct answer through straightforward reasoning. In contrast, Video-R1-7B experienced significant localization errors during the reasoning process, leading to a critical issue for reasoning models: a mismatch between the reasoning analysis and the final answer. LongVILA-R1-7B demonstrated superior performance by leveraging dense frame analysis to accurately infer the spatial relationships between rooms and furniture across different levels, ultimately delivering a coherent reasoning process and the correct answer. On Figure~\ref{fig:demo_temporal}, the focus shifts to temporal analysis in a Lego video featuring diverse events and interactions. All three models successfully reasoned through the sequence of events and provided correct answers, showcasing their proficiency in temporal reasoning tasks. As a supplement to Figure~\ref{fig:examples}, Figure~\ref{fig:demo_football} provides a more comprehensive comparison of two examples: "2022 FIFA Argentina vs. Netherlands" and "Moving the Cup and Guessing Where the Ball Is." In the football match example, while Gemini-1.5-Pro produced the correct answer, its output contained hallucinatory content influenced by biases in its pre-learned knowledge. Video-R1 not only failed to provide accurate video analysis reasoning but also made incorrect predictions. In contrast, LongVILA-R1 successfully analyzed the players' performance and emotions during the match, integrating these factors through its robust reasoning capabilities to make accurate predictions about the outcome. For the more challenging task of tracking the ball, Gemini-1.5-Pro's reasoning is inconsistent with the spatial content throughout, while Video-R1 failed to deduce the ball's final position accurately. Remarkably, LongVILA-R1 precisely analyzed the spatial transformations following the movement of the box, demonstrating superior interpretative and reasoning abilities.

\end{document}